\documentclass[11pt,letterpaper]{article}
\usepackage{naaclhlt2016}
\usepackage{times}
\usepackage{latexsym}
\usepackage{graphicx}

\newcommand\varlist{,\makebox[1em][c]{.\hfil.\hfil.},}

\naaclfinalcopy 
\def\naaclpaperid{514} 

\title{Multilingual Language Processing From Bytes}

\author{Dan Gillick, Cliff Brunk, Oriol Vinyals, Amarnag Subramanya \\
        Google Research \\
        \{dgillick, cliffbrunk, vinyals, asubram\}@google.com}

\date{}

\begin{document}

\maketitle

\begin{abstract}
We describe an LSTM-based model which we call Byte-to-Span (BTS) that reads
text as bytes and outputs span annotations of the form [start, length, label]
where start positions, lengths, and labels are separate entries in our
vocabulary.
Because we operate directly on unicode bytes rather than language-specific words or
characters, we can analyze text in many languages with a single model.
Due to the small vocabulary size, these multilingual models are
very compact, but produce results similar to or better than the
state-of-the-art in Part-of-Speech tagging and Named Entity Recognition that use
only the provided training datasets (no external data sources).
Our models are learning ``from scratch'' in that they do not rely on any
elements of the standard pipeline in Natural Language Processing (including tokenization), and thus can
run in standalone fashion on raw text.
\end{abstract}

\section{Introduction}

\begin{figure*}[htb]
\centering
\includegraphics[width=\linewidth]{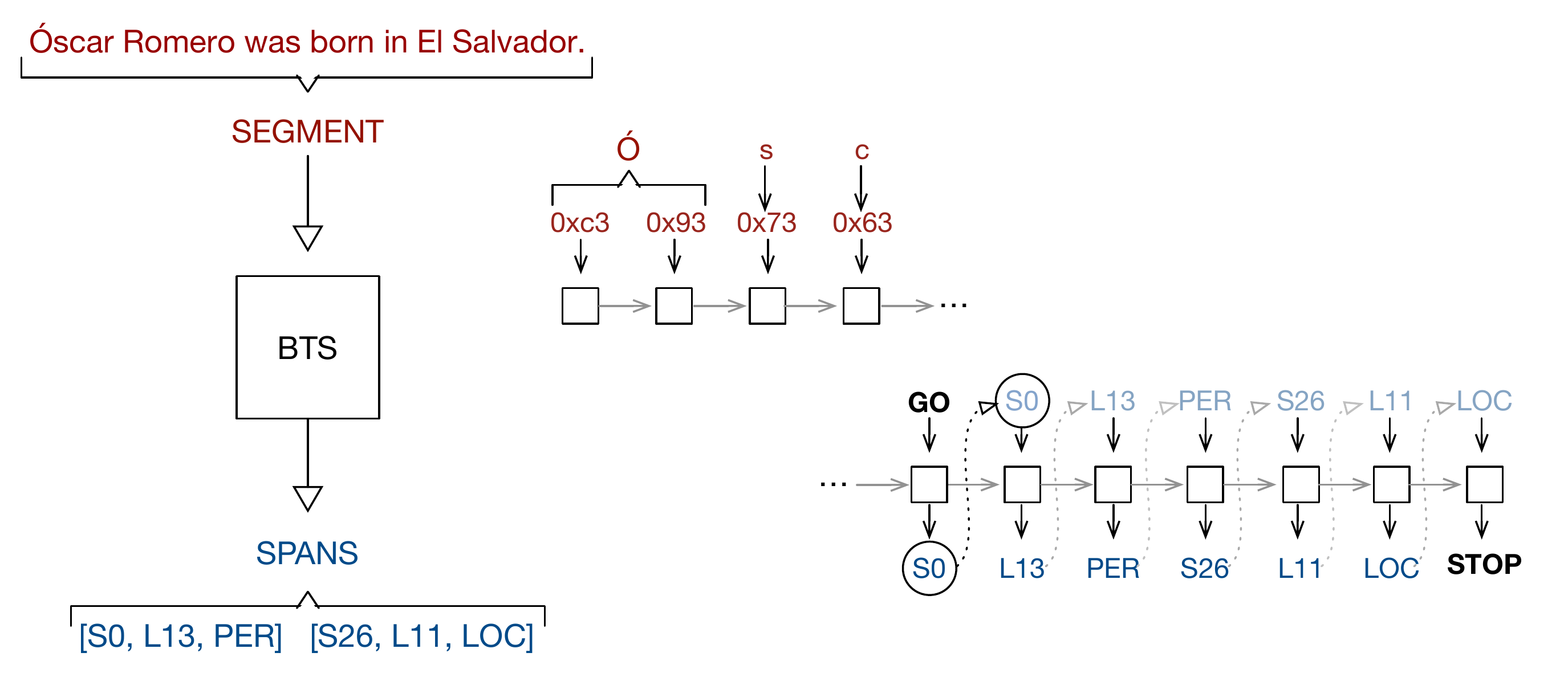}
\caption{\label{fig:bytes2spans} A diagram showing the way the
Byte-to-Span (BTS) model converts an input text segment to a sequence of
span annotations. The model reads the input segment one byte at a time
(this can involve multibyte unicode characters), then a special Generate Output (GO) symbol,
then produces the argmax output of a softmax over all possible start positions,
lengths, and labels (as well as STOP, signifying no additional outputs). The
prediction from the previous time step is fed as an input to the next time
step.}
\end{figure*}

The long-term trajectory of research in Natural Language Processing has seen the
replacement of rules and specific linguistic knowledge with machine learned
components.
Perhaps the most standardized way that knowledge is still injected into largely
statistical systems is through the processing pipeline:
Some set of basic language-specific tokens are identified in a first step.
Sequences of tokens are segmented into sentences in a second step.
The resulting sentences are fed one at a time for syntactic analysis:
Part-of-Speech (POS) tagging and parsing.
Next, the predicted syntactic structure is typically used as features in
semantic analysis, Named Entity Recognition (NER), Semantic Role Labeling, etc.
While each step of the pipeline now relies more on data and models than on
hand-curated rules, the pipeline structure itself encodes one particular
understanding of how meaning attaches to raw strings.

One motivation for our work is to try removing this structural dependence.
Rather than rely on the intermediate representations invented for specific
subtasks (for example, Penn Treebank tokenization), we are allowing the model to
learn whatever internal structure is most conducive to producing the
annotations of interest.
To this end, we describe a Recurrent Neural Network (RNN) model that reads
raw input string \textit{segments}, one byte at a time, and produces output
\textit{span annotations} corresponding to specific byte regions in the
input\footnote{Our span annotation model can be applied to
any sequence labeling task; it is not immediately applicable to predicting more
complex structures like trees.}.
This is truly language annotation from scratch
(see Collobert et al. \shortcite{collobert2011natural} and Zhang and
LeCun \shortcite{zhang2015text}).

Two key innovations facilitate this approach.
First, Long Short Term Memory (LSTM) models \cite{hochreiter1997long}
allow us to replace the traditional independence assumptions in text processing
with structural constraints on memory.
While we have long known that long-term dependencies are important in language,
we had no mechanism other than conditional independence to keep sparsity in
check. The memory in an LSTM, however, is not constrained by any explicit
assumptions of independence.
Rather, its ability to learn patterns is limited only by the structure of the
network and the size of the memory (and of course the amount of training data).

Second, sequence-to-sequence models \cite{sutskever2014sequence}, allow for
flexible input/output dynamics.
Traditional models, including feedforward neural networks, read fixed-length
inputs and generate fixed-length outputs by following a fixed set of
computational steps.
Instead, we can now read an entire segment of text before producing an
arbitrary number of outputs, allowing the model to learn a function best suited
to the task.

We leverage these two ideas with a basic strategy:
Decompose inputs and outputs into their component pieces, then read and predict
them as sequences.
Rather than read words, we are reading a sequence of unicode bytes\footnote{We
use the variable length UTF-8 encodings to keep the vocabulary as small as possible.};
rather than producing a label for each word, we are producing triples
[start, length, label], that correspond to the spans of interest, as a sequence
of three separate predictions (see Figure \ref{fig:bytes2spans}).
This forces the model to learn how the components of words and labels interact
so all the structure typically imposed by the NLP pipeline
(as well as the rules of unicode) are left to the LSTM to model.

Decomposed inputs and outputs have a few important benefits.
First, they reduce the size of the vocabulary relative to word-level inputs,
so the resulting models are extremely compact (on the order of a million
parameters).
Second, because unicode is essentially a universal language, we can train
models to analyze many languages at once.
In fact, by stacking LSTMs, we are able to learn representations that
appear to generalize across languages, improving performance significantly
(without using any additional parameters) over models trained on a single language.
This is the first account, to our knowledge, of a multilingual model that achieves
good results across many languages, thus bypassing all the
language-specific engineering usually required to
build models in different languages\footnote{These multilingual models are able
to handle code-mixed text, an important practical problem that's received relatively
little attention. However, we do not have any annotated data that contains
code switching, so we cannot report any results.}.
We describe results similar to or better than the
state-of-the-art in Part-of-Speech tagging and Named Entity Recognition that use
only the provided training datasets (no external data sources).

The rest of this paper is organized as follows.
Section 2 discusses related work;
Section 3 describes our model;
Section 4 gives training details including a new variety of dropout \cite{hinton2012improving};
Section 5 gives inference details;
Section 6 presents results on POS tagging and NER across many languages;
Finally, we summarize our contributions in section 7.

\section{Related Work}

One important feature of our work is the use of byte inputs.
Character-level inputs have been used with some success for tasks like NER
\cite{klein2003named}, parallel text alignment \cite{church1993char_align},
and authorship attribution \cite{peng2003language} as an effective way to deal
with n-gram sparsity while still capturing some aspects of word choice and
morphology.
Such approaches often combine character and word features and have been
especially useful for handling languages with large character sets
\cite{nakagawa2004chinese}.
However, there is almost no work that explicitly uses bytes -- one exception
uses byte n-grams to identify source code authorship
\cite{frantzeskou2006effective} -- but there is nothing, to the best of our
knowledge, that exploits bytes as a cross-lingual representation of language.
Work on multilingual parsing using Neural Networks that share some subset of the
parameters across languages \cite{duong2low} seems to benefit the low-resource
languages; however, we are sharing all the parameters among all languages.

Recent work has shown that modeling the sequence of characters in each
token with an LSTM can more effectively handle rare and unknown words than
independent word embeddings \cite{ling2015finding,ballesteros2015improved}.
Similarly, language modeling, especially for morphologically complex languages,
benefits from a Convolutional Neural Network (CNN) over characters to generate
word embeddings \cite{kim2015character}.
Rather than decompose words into characters, Rohan and Denero
\shortcite{chitnis2015variable} encode rare words with Huffman codes, allowing
a neural translation model to learn something about word subcomponents.
In contrast to this line of research, our work has no explicit notion of tokens
and operates on bytes rather than characters.

Our work is philosophically similar to Collobert et al.'s
\shortcite{collobert2011natural} experiments with ``almost from scratch''
language processing.
They avoid task-specific feature engineering, instead relying on a multilayer
feedforward (or convolutional) Neural Network to combine word embeddings to
produce features useful for each task.
In the Results section, below, we compare NER performance on the same dataset
they used.
The ``almost'' in the title actually refers to the use of preprocessed
(lowercased) tokens as input instead of raw sequences of letters.
Our byte-level models can be seen as a realization of their comment: ``A completely
from scratch approach would presumably not know anything about words at all and
would work from letters only.''
Recent work with convolutional neural networks that read character-level
inputs \cite{zhang2015text} shows some interesting
results on a variety of classification tasks, but because their models need
very large training sets, they do not present
comparisons to established baselines on standard tasks.

Finally, recent work on Automatic Speech Recognition (ASR) uses a similar
sequence-to-sequence LSTM framework to produce letter sequences directly from
acoustic frame sequences \cite{chan2015listen,bahdanau2015end}.
Just as we are discarding the usual intermediate representations used for text
processing, their models make no use of phonetic alignments, clustered
triphones, or pronunciation dictionaries.
This line of work -- discarding intermediate representations in speech -- was
pioneered by Graves and Jaitly \shortcite{graves2014towards} and earlier, by
Eyben et al. \shortcite{eyben2009speech}.

\section{Model}

Our model is based on the sequence-to-sequence model used for machine
translation \cite{sutskever2014sequence}, an adaptation of an LSTM that
encodes a variable length input as a fixed-length vector, then decodes it into a
variable number of outputs\footnote{Related translation work adds an attention
mechanism \cite{bahdanau2014neural}, allowing the decoder to attend directly to
particularly relevant inputs. We tried adding the same mechanism to our model but saw no
improvement in performance on the NER task, though training converged in fewer
steps.}.

Generally, the sequence-to-sequence LSTM is trained to estimate the conditional probability
$P(y_1\varlist y_{T'}|x_1\varlist x_T)$ where $(x_1\varlist x_T)$ is an input
sequence and $(y_1\varlist y_{T'})$ is the corresponding output sequence whose
length $T'$ may differ from $T$.
The encoding step computes a fixed-dimensional representation $v$ of
the input $(x_1\varlist x_T)$ given by the hidden state of the LSTM after
reading the last input $x_T$.
The decoding step computes the output probability $P(y_1\varlist y_{T'})$ with the
standard LSTM formulation for language modeling, except that the initial hidden
state is set to $v$:

\begin{equation}
P(y_1\varlist y_{T'}|x_1\varlist x_T) = \prod_{t=1}^{T'} P(y_t|v,y_1\varlist y_{t-1})
\end{equation}

Sutskever et al. used a separate LSTM for the encoding and decoding tasks.
While this separation permits training the encoder and decoder LSTMs separately,
say for multitask learning or pre-training, we found our results were no
worse if we used a single set of LSTM parameters for both encoder and decoder.

\subsection{Vocabulary}

The primary difference between our model and the translation model is our
novel choice of vocabulary.
The set of inputs include all 256 possible bytes, a special Generate Output (GO)
symbol, and a special DROP symbol used for regularization, which we will discuss
below.
The set of outputs include all possible span start positions (byte $0..k$), all
possible span lengths ($0..k$), all span labels ({PER, LOC, ORG, MISC} for
the NER task), as well as a special STOP symbol.
A complete span annotation includes a start, a length, and a label, but as
shown in Figure \ref{fig:bytes2spans}, the model is trained to produce this triple as
three separate outputs.
This keeps the vocabulary size small and in practice, gives better performance
(and faster convergence) than if we use the cross-product space of the
triples.

More precisely, the prediction at time $t$ is conditioned on the full
input and all previous predictions (via the chain rule).
By splitting each span annotation into a sequence [start, length, label],
we are making no independence assumption; instead we are relying on
the model to maintain a memory state that captures the important dependencies.

Each output distribution $P(y_t|v,y_1\varlist y_{t-1})$ is given by a softmax
over all possible items in the output vocabulary, so at a given time step,
the model is free to predict any start, any length, or any label (including
STOP).
In practice, because the training data always has these complete triples in a
fixed order, we seldom see malformed or incomplete spans (the decoder
simply ignores such spans).
During training, the true label $y_{t-1}$ is fed as input to the model
at step $t$ (see Figure \ref{fig:bytes2spans}), and during inference, the
argmax prediction is used instead.
Note also that the training procedure tries to maximize the probability in
Equation 1 (summed over all the training examples).
While this does not quite match our task objectives (F1 over labels, for
example), it is a reasonable proxy.

\subsection{Independent segments}

Ideally, we would like our input segments to cover full documents so that our
predictions are conditioned on as much relevant information as possible.
However, this is impractical for a few reasons.
From a training perspective, a Recurrent Neural Network is unrolled to resemble
a deep feedforward network, with each layer corresponding to a time step.
It is well-known that running backpropagation over a very deep network is
hard because it becomes increasingly difficult to estimate the contribution of
each layer to the gradient, and further, RNNs have trouble generalizing to
different length inputs \cite{erhan2009difficulty}.

So instead of document-sized input segments, we make a segment-independence
assumption:
We choose some fixed length $k$ and train the model on segments of length $k$
(any span annotation not completely contained in a segment is ignored).
This has the added benefit of limiting the range of the start and length label
components.
It can also allow for more efficient batched inference since each segment is
decoded independently.
Finally, we can generate a large number of training segments by sliding a window
of size $k$ one byte at a time through a document.
Note that the resulting training
segments can begin and end mid-word, and indeed, mid-character.
For both tasks described below, we set the segment size $k=60$.

\subsection{Sequence ordering}

Our model differs from the translation model in one more important way.
Sutskever et al. found that feeding the input words in reverse order and generating the output words in
forward order gave significantly better translations, especially for long sentences.
In theory, the predictions are conditioned on the entire input, but as a
practical matter, the learning problem is easier when relevant information is
ordered appropriately since long dependencies are harder to learn than short
ones.

Because the byte order is more meaningful in the forward direction (the first
byte of a multibyte character specifies the length, for example), we found
somewhat better performance with forward order than reverse order (less than 1\% absolute).
But unlike translation, where the outputs have a complex order determined by the
syntax of the language, our span annotations are more like an unordered set.
We tried sorting them by end position in both forward and backward order, and
found a small improvement (again, less than 1\% absolute) using the backward ordering (assuming the input
is given in the forward order).
This result validates the translation ordering experiments: the modeling
problem is easier when the sequence-to-sequence LSTM is used more like a
stack than a queue.

\subsection{Model shape}

We experimented with a few different architectures and found no
significant improvements in using more than 320 units for the embedding
dimension and LSTM memory and 4 stacked LSTMs (see Table \ref{tab:width-depth}).
This observation holds for both models trained on a single language and models
trained on many languages.
Because the vocabulary is so small, the total number of parameters is dominated
by the size of the recurrent matrices.
All the results reported below use the same architecture (unless otherwise noted)
and thus have roughly 900k parameters.

\section{Training}

We trained our models with Stochastic Gradient Descent (SGD) on mini-batches of
size 128, using an initial learning rate of 0.3.
For all other hyperparameter choices, including random initialization,
learning rate decay, and gradient clipping, we follow Sutskever et
al. \shortcite{sutskever2014sequence}.
Each model is trained on a single CPU over a period of a few days, at which
point, development set results have stabilized.
Distributed training on GPUs would likely speed up training to just a few hours.

\subsection{Dropout and byte-dropout}

Neural Network models are often trained using \textit{dropout}
\cite{hinton2012improving}, which tends to improve generalization by limiting
correlations among hidden units.
During training, dropout randomly zeroes some fraction of the elements
in the embedding layer and the model state just before the softmax layer
\cite{zaremba2014recurrent}.

We were able to further improve generalization with a technique we are calling
\textit{byte-dropout}:
We randomly replace some fraction of the input bytes in each segment with a
special DROP symbol (without changing the corresponding span annotations).
Intuitively, this results in a more robust model, perhaps by forcing it to use
longer-range dependencies rather than memorizing particular local sequences.

It is worth noting that noise is often added at training time to images in image classification
and speech in speech recognition where the added noise does not fundamentally
alter the input, but rather blurs it.
By using a byte representation of language, we are now capable of achieving something
like blurring with text.
Indeed, if we removed 20\% of the characters in a sentence, humans would be
able to infer words and meaning reasonably well.

\section{Inference}

We perform inference on a segment by (greedily) computing the most likely output
at each time step and feeding it to the next time step.
Experiments with beam search show no meaningful improvements (less than 0.2\% absolute).
Because we assume that each segment is independent, we need to choose how to
break up the input into segments and how to stitch together the results.

The simplest approach is to divide up the input into segments with no
overlapping bytes.
Because the model is trained to ignore incomplete spans, this approach misses
all spans that cross segment boundaries, which, depending on the choice of $k$,
can be a significant number. We avoid the missed-span problem by choosing
segments that overlap such that each span is likely to be fully contained by at least one
segment.

For our experiments, we create segments with a fixed overlap ($k/2=30$).
This means that with the exception of the first segment in a document,
the model reads 60 bytes of input, but we only keep predictions about the last 30 bytes.

\section{Results}

Here we describe experiments on two datasets that include annotations across
a variety of languages.
The multilingual datasets allow us to highlight the advantages of using
byte-level inputs:
First, we can train a single compact model that can handle many languages at
once.
Second, we demonstrate some cross-lingual abstraction that improves
performance of a single multilingual model over each single-language model.
In the experiments, we refer to the LSTM setup described above as Byte-to-Span
or BTS.

Most state-of-the-art results in POS tagging and NER
leverage unlabeled data to improve a supervised baseline.
For example, word clusters or word embeddings estimated from a large corpus are
often used to help deal with sparsity.
Because our LSTM models are reading bytes, it is not obvious how to insert
information like a word cluster identity.
Recent results with sequence-to-sequence auto-encoding \cite{dai2015semi} seem
promising in this regard, but here we limit our experiments to use
just annotated data.

Each task specifies separate data for training, development, and testing.
We used the development data for tuning the dropout and byte-dropout parameters
(since these likely depend on the amount of available training data),
but did not tune the remaining hyperparameters.
In total, our training set for POS Tagging across 13 languages included 2.87
million tokens and our training set for NER across 4 languages included 0.88
million tokens.
Recall, though, that our training examples are 60-byte segments
obtained by sliding a window through the training data, shifting by 1 byte each
time.
This results in 25.3 million and 6.0 million training segments for the two tasks.

\subsection{Part-of-Speech Tagging}

Our part-of-speech tagging experiments use Version 1.1 of the Universal Dependency
data\footnote{http://universaldependencies.github.io/docs/}, a collection of
treebanks across many languages annotated with a universal tagset
\cite{petrov2011universal}.
The most relevant recent work \cite{ling2015finding} uses different datasets,
with different finer-grained tagsets in each language.
Because we are primary interested in multilingual models that can share
language-independent parameters, the universal tagset is important, and thus
our results are not immediately comparable.
However, we provide baseline results (for each language separately) using a
Conditional Random Field \cite{lafferty2001conditional} with an extensive
collection of features with performance comparable to the Stanford POS tagger
\cite{manning2011part}.
For our experiments, we chose the 13 languages that had at least 50k tokens of
training data.
We did not subsample the training data, though the amount of data varies widely
across languages, but rather shuffled all training examples together.
These languages represent a broad range of linguistic phenomena and character
sets so it was not obvious at the outset that a single multilingual model would
work.

Table \ref{tab:pos-by-language} compares the baselines with (CRF+) and without (CRF) externally
trained cluster features with our model trained on all languages (BTS) as well as each
language separately (BTS*).
The single BTS model improves on average over the CRF models trained using the
same data, though clearly there is some benefit in using external resources.
Note that BTS is particularly strong in Finnish, surpassing even CRF+ by
nearly 1.5\% (absolute), probably because the byte representation generalizes
better to agglutinative languages than word-based models, a finding validated by
Ling et al. \shortcite{ling2015finding}.
In addition, the baseline CRF models, including the (compressed) cluster tables,
require about 50 MB per language, while BTS is under 10 MB.
BTS improves on average over BTS*, suggesting that it is learning some language-independent
representation.

\begin{table}[htb]
\centering
\begin{tabular}{|l|cc|cc|}
\hline
\bf Language & \bf CRF+ & \bf CRF & \bf BTS & \bf BTS* \\
\hline
Bulgarian  & 97.97 & 97.00 & 97.84 & 97.02 \\
Czech      & 98.38 & 98.00 & 98.50 & 98.44 \\
Danish     & 95.93 & 95.06 & 95.52 & 92.45 \\
German     & 93.08 & 91.99 & 92.87 & 92.34 \\
Greek      & 97.72 & 97.21 & 97.39 & 96.64 \\
English    & 95.11 & 94.51 & 93.87 & 94.00 \\
Spanish    & 96.08 & 95.03 & 95.80 & 95.26 \\
Farsi      & 96.59 & 96.25 & 96.82 & 96.76 \\
Finnish    & 94.34 & 92.82 & 95.48 & 96.05 \\
French     & 96.00 & 95.93 & 95.75 & 95.17 \\
Indonesian & 92.84 & 92.71 & 92.85 & 91.03 \\
Italian    & 97.70 & 97.61 & 97.56 & 97.40 \\
Swedish    & 96.81 & 96.15 & 95.57 & 93.17 \\
\hline
AVERAGE    & 96.04 & 95.41 & 95.85 & 95.06 \\
\hline
\end{tabular}
\caption{\label{tab:pos-by-language} Part-of-speech tagging accuracy for two CRF
baselines and 2 versions of BTS.
CRF+ uses resources external to the training data (word clusters) and CRF uses
only the training data.
BTS (unlike CRF+ and CRF) is a single model trained on all the languages together,
while BTS* is a separate Byte-to-Span model for each language.}
\end{table}

\subsection{Named Entity Recognition}

Our main motivation for showing POS tagging results was to demonstrate how
effective a single BTS model can be across a wide range of languages.
The NER task is a more interesting test case because, as discussed in the
introduction, it usually relies on a pipeline of processing.
We use the 2002 and 2003 ConLL shared task
datasets\footnote{http://www.cnts.ua.ac.be/conll200\{2,3\}/ner}
for multilingual NER because they contain data in 4 languages
(English, German, Spanish, and Dutch) with consistent annotations of named
entities (PER, LOC, ORG, and MISC).
In addition, the shared task competition produced strong baseline numbers for
comparison.
However, most published results use extra information beyond the provided
training data which makes fair comparison with our model more difficult.

The best competition results for English and German \cite{florian2003named} used a large gazetteer and the
output of two additional NER classifiers trained on richer datasets.
Since 2003, better results have been reported using additional
semi-supervised techniques \cite{ando2005framework} and more recently,
Passos et al. \shortcite{passos2014lexicon} claimed the best English results
(90.90\% F1) using features derived from word-embeddings.
The 1st place submission in 2002 \cite{carreras2002conll} comment that without
extra resources for Spanish, their results drop by about 2\% (absolute).

Perhaps the most relevant comparison is the overall 2nd place submission in 2003
\cite{klein2003named}.
They use only the provided data and report results with character-based
models which provide a useful comparison point to our byte-based LSTM.
The performance of a character HMM alone is much worse than their best
result (83.2\% vs 92.3\% on the English development data),
which includes a variety of word and POS-tag features that describe the
context (as well as some post-processing rules).
For English (assuming just ASCII strings), the character HMM uses the same
inputs as BTS, but is hindered by some combination of the independence
assumption and smaller capacity.

Collobert et al.'s \shortcite{collobert2011natural} convolutional model
(discussed above) gives 81.47\% F1 on the English test set when trained on only
the gold data.
However, by using carefully selected word-embeddings trained on external data,
they are able to increase F1 to 88.67\%.
Huang et al. \shortcite{huang2015bidirectional} improve on Collobert's results
by using a bidirectional LSTM with a CRF layer
where the inputs are features describing the words in each sentence.
Either by virtue of the more powerful model, or because of more expressive
features, they report 84.26\% F1 on the same test set and 90.10\% when they add
pretrained word embedding features.
Dos Santos et al. \shortcite{dos2015boosting} represent each word by
concatenating a pretrained word embedding with a character-level embedding
produced by a convolutional neural network.

There is relatively little work on multilingual NER, and most research is
focused on building systems that are unsupervised in the sense that they use
resources like Wikipedia and Freebase rather than manually annotated data.
Nothman et al. \shortcite{nothman2013learning} use Wikipedia anchor links and
disambiguation pages joined with Freebase types to create a huge
amount of somewhat noisy training data and are able to achieve good results
on many languages (with some extra heuristics).
These results are also included in Table \ref{tab:ner}.

\begin{table}[htb]
\centering
\begin{tabular}{|l|c|c|c|c|}
\hline
\bf Model & \bf en & \bf de & \bf es & \bf nl \\
\hline
Passos    & 90.90 & --    & --    & --    \\
Ando      & 89.31 & 75.27 & --    & --    \\
Florian   & 88.76 & 72.41 & --    & --    \\
Carreras  & --    & --    & 81.39 & 77.05 \\
dos Santos & --   & --    & 82.21 & --    \\
Nothman   & 85.2~~ & 66.5~~ & 79.6~~  & 78.6~~  \\
\hline
Klein     & 86.07 & 71.90 & --    & --    \\
Huang     & 84.26 & --    & --    & --    \\
Collobert & 81.47 & --    & --    & --    \\
\hline
BTS       & 86.50 & 76.22 & 82.95 & 82.84 \\
BTS*      & 84.57 & 72.08 & 81.83 & 78.08 \\
\hline
\end{tabular}
\caption{\label{tab:ner} A comparison of NER systems. The results are F1 scores,
where a correct span annotation exactly matches a gold span annotation (start, length, and
entity type must all be correct). Results of the systems described in the text
are shown for English, German, Spanish, and Dutch. BTS* shows the results of the
BTS model trained separately on each language while BTS is a single model trained
on all 4 languages together. The top set of results leverage resources beyond
the training data; the middle set do not, and thus are most comparable to our
results (bottom set).}
\end{table}

While BTS does not improve on the state-of-the-art in English, its performance
is better than the best previous results that use only the provided training data.
BTS improves significantly on the best known results in German, Spanish, and Dutch even
though these leverage external data.
In addition, the BTS* models, trained separately on each language, are worse than
the single BTS model (with the same number of parameters as each single-language
model) trained on all languages combined, again suggesting that
the model is learning some language-independent representation of the task.

One interesting shortcoming of the BTS model is that it is not obvious how to
tune it to increase recall.
In a standard classifier framework, we could simply increase the prediction
threshold to increase precision and decrease the prediction threshold to
increase recall.
However, because we only produce annotations for spans (non-spans are not
annotated), we can adjust a threshold on total span probability
(the product of the start, length, and label probabilities) to increase
precision, but there is no clear way to increase recall.
The untuned model tends to prefer precision over recall already, so some
heuristic for increasing recall might improve our overall F1 results.

\subsection{Dropout and Stacked LSTMs}

There are many modeling options and hyperparameters that significantly impact
the performance of Neural Networks.
Here we show the results of a few experiments that were particularly relevant
to the performance obtained above.

First, Table \ref{tab:pos-avg} shows how dropout and
byte-dropout improve performance for both tasks.
Without any kind of dropout, the training process starts to overfit
(development data perplexity starts increasing) relatively quickly.
For POS tagging, we set dropout and byte-dropout to 0.2, while for NER, we set
both to 0.3.
This significantly reduces the overfitting problem.


\begin{table}[htb]
\centering
\begin{tabular}{|l|c|c|}
\hline
\bf BTS Training & \bf POS Accuracy & \bf NER F1 \\
\hline
Vanilla         & 94.78 & 74.75 \\
+ Dropout       & 95.35 & 78.76 \\
+ Byte-dropout  & 95.85 & 82.13 \\
\hline
\end{tabular}
\caption{\label{tab:pos-avg} BTS Part-of-speech tagging average accuracy across
all 13 evaluated languages and Named Entity Recognition average F1 across all 4
evaluated languages with various modifications to the vanilla training setup.
Dropout is standard in Neural Network model training because it often improves
generalization; Byte-dropout randomly replaces input bytes with a special
DROP marker.}
\end{table}

\begin{table}[htb]
\centering
\begin{tabular}{|c|c|c|}
\hline
\bf Depth & \bf Width=320 & \bf Width=640 \\
\hline
1 & 76.15 & 77.59 \\
2 & 79.40 & 79.73 \\
3 & 81.44 & 81.93 \\
4 & 82.13 & 82.18 \\
\hline
\end{tabular}
\caption{\label{tab:width-depth} Macro-averaged (across 4 languages) F1 for the
NER task using different model architectures.}
\end{table}

Second, Table \ref{tab:width-depth} shows how performance improves as we increase
the size of the model in two ways: the number of units in the model's state
(width) and the number of stacked LSTMs (depth).
Increasing the width of the model improves performance less than increasing the
depth, and once we use 4 stacked LSTMs, the added benefit of a much wider model
has disappeared.
This result suggests that rather than learning to partition the space of inputs
according to the source language, the model is learning some lanugage-independent
representation at the deeper levels.

\begin{figure*}[htb]
\centering
\includegraphics[width=\linewidth]{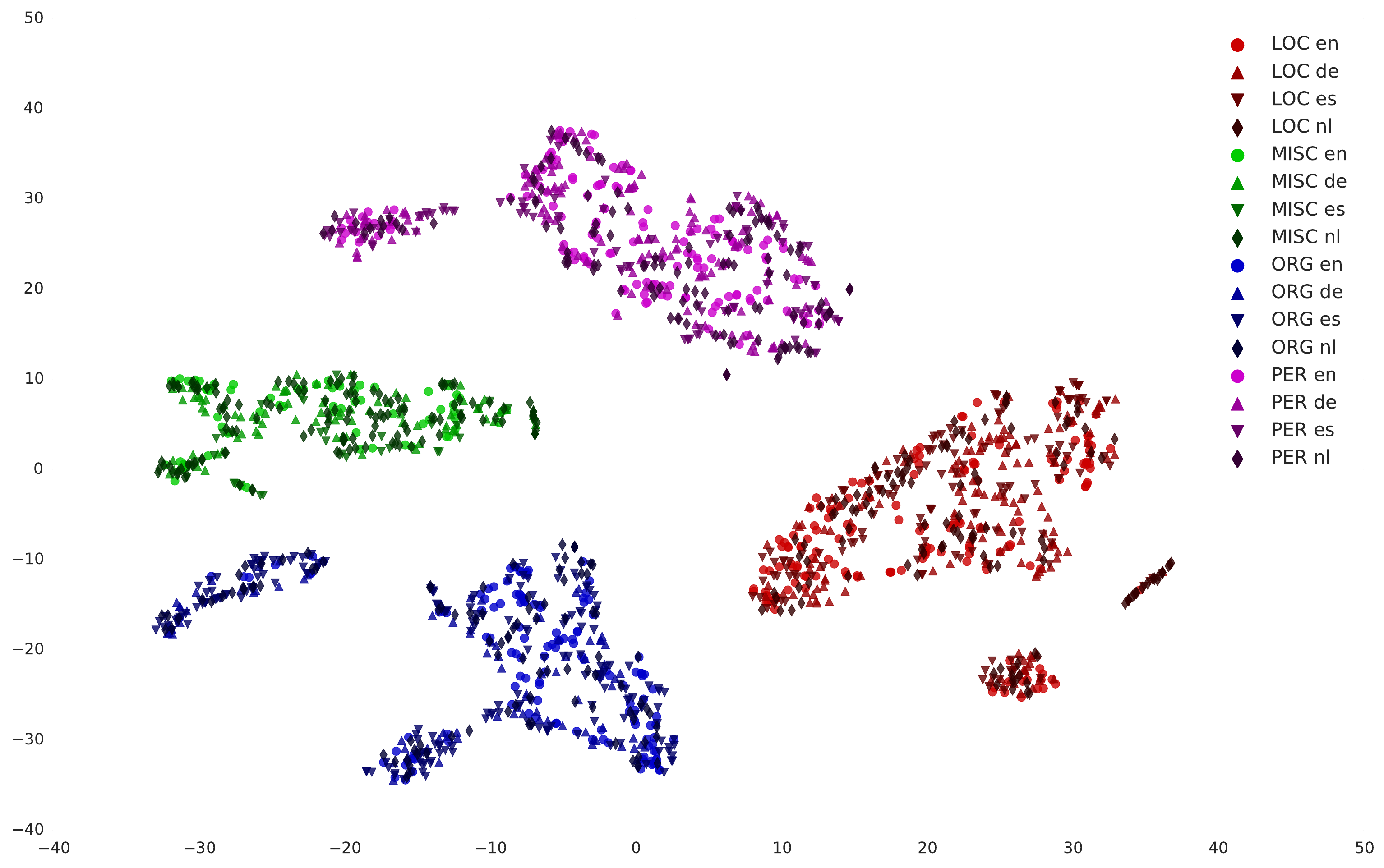}
\caption{\label{fig:tsne} A tSNE plot of the BTS model's memory state just before
the softmax layer produces one of the NER labels.}
\end{figure*}

To validate our claim about language-independent representation, Figure
\ref{fig:tsne} shows the results of a tSNE plot of the LSTM's memory state
when the output is one of {PER, LOC, ORG, MISC} across the four languages.
While the label clusters are neatly separated, the examples of each individual
label do not appear to be clustered by language.
Thus rather than partitioning each (label, language) combination, the model is
learning unified label representations that are independent of the language.

\section{Conclusions}
We have described a model that uses a sequence-to-sequence LSTM framework that
reads a segment of text one byte at a time and then produces span annotations
over the inputs.
This work makes a number of novel contributions:

First, we use the bytes in variable length unicode encodings as inputs. This
makes the model vocabulary very small and also allows us to train a multilingual
model that improves over single-language models without using additional
parameters. We introduce byte-dropout, an analog to added noise in speech or
blurring in images, which significantly improves generalization.

Second, the model produces span annotations, where each is a sequence of three
outputs: a start position, a length, and a label. This decomposition keeps the
output vocabulary small and marks a significant departure from the typical
Begin-Inside-Outside (BIO) scheme used for labeling sequences.

Finally, the models are much more compact than traditional word-based systems and
they are standalone -- no processing pipeline is needed. In particular, we do not
need a tokenizer to segment text in each of the input languages.

\section*{Acknowledgments}
Many thanks to Fernando Pereira and Dan Ramage for their insights about this
project from the outset. Thanks also to Cree Howard for creating Figure 1.

\bibliography{bytelstm}
\bibliographystyle{naaclhlt2016}

\end{document}